\documentclass{article}

\usepackage[preprint,dblblindworkshop,nonatbib]{neurips_2026}
\workshoptitle{Real-Time Conversational Agents}

\usepackage[T1]{fontenc}
\usepackage[utf8]{inputenc}
\usepackage{microtype}
\usepackage{booktabs}
\usepackage{graphicx}
\usepackage{amsmath}
\usepackage{amssymb}
\usepackage[table]{xcolor}
\usepackage{caption}
\usepackage{tikz}
\usetikzlibrary{arrows.meta,positioning}
\usepackage{listings}
\usepackage{url}

\definecolor{cangry}{HTML}{D62728}
\definecolor{chappy}{HTML}{D4A017}
\definecolor{csad}{HTML}{1F77B4}
\definecolor{csurprise}{HTML}{8C564B}
\definecolor{cneutral}{HTML}{7F7F7F}

\title{Steering Follows Geometry, Not Labels: Emotion Directions in a Full-Duplex Speech Model}
\author{%
  Pulak Kuli\\
  Independent Researcher\\
  \texttt{pulakkuli72@gmail.com}
}

\begin{document}
\maketitle

\begin{abstract}
Full-duplex voice agents need to modulate emotion and delivery during real-time conversations, when de-escalating a
complaint, carrying urgency in dispatch, softening a clinical result. Emotion and delivery control
is well studied for TTS and turn based models through prompt-conditioned synthesis,
reference-conditioned synthesis and activation steering; PersonaPlex~[10] controls identity in a
duplex model but not affect.

We study emotion steering in Moshi, a fully open sourced full-duplex speech language model, across four emotions, using mean-difference activation steering,
which costs only a few vector additions per frame and no retraining. 

We show that emotion is linearly decodable from Moshi's
residual stream, but activation steering is only partially achievable, and unevenly so; as happy, angry and surprise steer towards a shared direction while sad is distinctly steerable. We also show that the shared component across the three emotions cannot simply be projected away from all the emotions equally.
\end{abstract}

\section{Introduction}
Emotion and delivery control is a well understood problem statement for voice agents with use-cases spanning across various real time voice interactions involving de-escalating a complaint by modulating delivery, carrying urgency in dispatch, or softening clinical results. Different scenarios require varying degrees of emotion control in voice agents which humans are able to gauge and dictate quite effectively.

Mechanisms such as prompt-conditioned and reference-audio-conditioned audio synthesis have been studied extensively [1,2], which suits turn based or Text-to-Speech (TTS) synthesizers. Activation steering [3,4] is another effective and efficient methodology to control emotion or delivery, as shown in EmoSteer-TTS [5]. Full-duplex speech models are superseding turn-based voice agents due to improved natural communication with humans.

With Full duplex models becoming more powerful and feasible in the modern world, the effect of activation steering becomes more relevant in the context of emotion and delivery control. At the same time prior work understudies its impact on newer full-duplex speech language models such as Moshi [6], which listens and speaks at once, emits audio frame by frame at 12.5\,Hz through a neural codec, and carries its own dialogue state. Recent full duplex models, such as PersonaPlex~[10], which is an extension to Moshi, focus primarily on identity control via prompt conditioning, instead of dynamic delivery or emotion control.

Whether a mean-difference vector direction still
controls emotion effectively, under those conditions is untested. In this paper we analyze Moshi's[6] capability to steer towards four emotions; angry, sad, happy and surprise. 

Our contributions include:

\textbf{(i)} A linear probe that is able to read the emotions accurately off the residual stream (\S\ref{sec:decode}).

\textbf{(ii)} A diagnosis about the effectiveness of mean-difference activation steering for all the four emotions (\S\ref{sec:geom})

\textbf{(iii)} A demonstration of shared common axis across happy, angry and surprise emotions that explain the gaps, with their cosine similarity being very high compared to a cross-speaker variance of the residual stream.

\textbf{(iv)} A method to project out the shared component to fix angry, but not as effectively for surprise. (\S\ref{sec:sweep})

\textbf{(v)} An analysis of the model's own generative limitations to produce emotions in audio (\S\ref{sec:attribution}).

\section{Related work}

\textbf{Activation steering for emotion in full duplex models.} Adding a direction to hidden states controls behaviour in
text language models [3,4]. EmoSteer-TTS [5] brings this to speech, applying difference vectors to a
searched subset of activations in flow-matching synthesisers, training-free. Two recent studies
question whether a single direction is the right instrument at all. Du et al.~[15] find with sparse
autoencoders that emotional variation in an LLM-based TTS model is distributed over several latent
features rather than one global shift, and that intervening on a chosen subset outperforms global
steering. Wang et al.~[16] compare steering sites geometrically and report that a flow-matching
module generalises poorly across speakers because speaker and emotion are entangled there, while the
speech-LM module offers a cleaner low-dimensional emotion subspace. Braun et al.~[17,18] make the same
kind of geometric argument in text language models, predicting steering reliability from the cosine
structure of the activation differences a vector is built from - measured within one behaviour, where
ours is measured between emotions.

Our result is the duplex
counterpart of both observations.

Steering has reached full-duplex models, though not for delivery:
Chang et al.~[12] add a perception vector to reduce the \emph{state inertia} that makes a model miss
the opening of an interruption, training-free, on Moshi among others. Duplex systems that do control
delivery train for it - F-Actor~[13] conditions on voice, topic and conversational behaviour, and
JoyAI-Talker~[14] takes delivery instructions at inference after fine-tuning on expressive speech. We
are not aware of prior work controlling emotion in a streaming duplex model by editing activations.

\section{Method}
\label{sec:method}

\subsection{Construction of emotion direction vectors}

We write real ESD~[7] recordings into Moshi's output codebooks, with their transcripts forced through Moshi's inner monologue, and let it run a forward pass. Moshi's audio output is a stream of Mimi codebook tokens that can be overridden. For those frames the model is producing that recording.

We read the residual stream over the last few frames before the override ends, by which point the
model has been carried through the whole utterance rather than its opening. Averaging over frames and
clips and subtracting the same quantity on neutral recordings gives one direction per layer,
\begin{equation}
v_l = \mathrm{mean}(h_l \mid e) - \mathrm{mean}(h_l \mid \mathrm{neutral}),
\end{equation}
which we unit-normalise. Whether that state is genuinely emotional is measured at \S\ref{sec:attribution}.

\subsection{Injecting direction vectors to steer emotion}
\label{sec:inject}

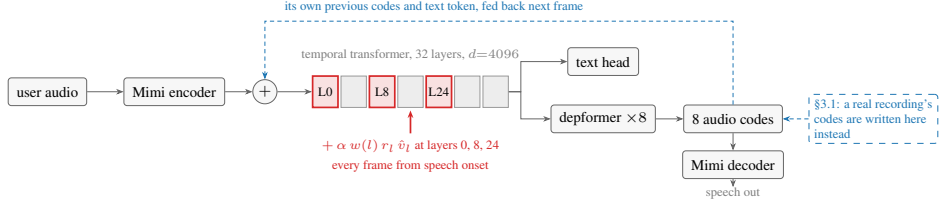
\begin{figure}[!htb]
\centering
\resizebox{0.88\linewidth}{!}{%
\begin{tikzpicture}[
  x=1cm, y=1cm, font=\scriptsize,
  box/.style={draw=black!60, rounded corners=1.5pt, minimum height=0.46cm, inner xsep=4pt,
              align=center, fill=black!3},
  lay/.style={draw=black!40, fill=black!7, minimum width=0.42cm, minimum height=0.52cm,
              inner sep=0pt},
  hit/.style={draw=cangry, line width=0.8pt, fill=cangry!14, minimum width=0.42cm,
              minimum height=0.52cm, inner sep=0pt, font=\tiny},
  flow/.style={-{Stealth[length=3.4pt]}, black!60},
  loop/.style={-{Stealth[length=3.4pt]}, csad, dashed, dash pattern=on 2pt off 1.5pt},
  inject/.style={-{Stealth[length=3.6pt]}, cangry, line width=0.8pt}
]

\node[box] (audio) at (0.95, 1.55) {user audio};
\node[box] (enc)   at (3.05, 1.55) {Mimi encoder};
\node[circle, draw=black!60, fill=black!3, inner sep=1.2pt] (sum) at (4.55, 1.55) {$+$};

\foreach \x/\name/\id in {5.55/L0/nA, 6.02//nB, 6.49/L8/nC, 6.96//nD,
                          7.43/L24/nE, 7.90//nF, 8.37//nG}
  \node[\ifx\name\empty lay\else hit\fi] (\id) at (\x, 1.55) {\name};
\node[font=\tiny, black!50, align=center] at (6.96, 2.16)
  {temporal transformer, 32 layers, $d{=}4096$};

\draw[inject] (6.96, 0.86) -- (6.96, 1.24);
\node[cangry, font=\tiny, align=center, anchor=north, inner sep=2pt] at (6.96, 0.84)
  {$+\,\alpha\,w(l)\,r_l\,\hat v_l$ at layers 0, 8, 24\\[2.5pt] every frame from speech onset};

\node[box] (text) at (10.15, 2.05) {text head};
\node[box] (dep)  at (10.15, 1.10) {depformer $\times 8$};
\node[box] (codes) at (12.30, 1.10) {8 audio codes};
\node[box] (dec)  at (12.30, 0.34) {Mimi decoder};
\node[font=\tiny, black!50] at (12.30, -0.13) {speech out};

\draw[flow] (audio) -- (enc);
\draw[flow] (enc) -- (sum);
\draw[flow] (sum) -- (nA.west);
\draw[flow] (nG.east) -- ++(0.20,0) |- (text.west);
\draw[flow] (nG.east) -- ++(0.20,0) |- (dep.west);
\draw[flow] (dep) -- (codes);
\draw[flow] (codes) -- (dec);
\draw[flow] (dec.south) -- (12.30, -0.03);

\draw[loop] (codes.north) |- (10.15, 2.72) -- (4.55, 2.72) -- (sum.north);
\node[csad, font=\tiny, anchor=south] at (7.35, 2.74)
  {its own previous codes and text token, fed back next frame};

\node[csad, draw=csad, dashed, rounded corners=1.5pt, font=\tiny, align=left,
      inner sep=2.5pt, anchor=west] (ovr) at (13.55, 1.10)
  {\S3.1: a real recording's\\ codes are written here\\ instead};
\draw[loop] (ovr.west) -- (codes.east);
\end{tikzpicture}}
\caption{Where the vector enters. Moshi is frozen; the only change is one direction added to the
output of the three layers marked in red - 0, 8 and 24 - every frame from speech onset. The injection sits upstream of the
split into the text head and the audio depformer, so it can shape delivery without touching the
codec.}
\label{fig:arch}
\end{figure}

At the injection stage, we interpolate the output of layers 0, 8 and
24, at every frame (Figure~\ref{fig:arch}) towards the desired emotion, by adding $\alpha \cdot w(l) \cdot r_l \cdot \hat{v}_l$. Here $r_l$ is that layer's median residual
norm, so $\alpha$ is relative to the activations it is added to, and $w(l) = 1 - l/31$ weights deeper
layers down. The depformer that predicts the audio codebooks, the codec and the other 29 layers are left
untouched.

We use $\alpha = 0.30$ and this three-layer schedule throughout the experiments. Both were chosen in preliminary work
on held-out sentences: strength peaks at $0.30$ and falls away past it as intelligibility degrades
faster than the emotion improves, and spreading the push across layers 0, 8 and 24 beat concentrating
it at specific layers.

\subsection{Probing the residual stream}

To determine whether Moshi is able to distinguish the emotions accurately, we fit one
multinomial logistic regression per layer to 60 clips per emotion drawn from the same states. 

Inputs are
standardised and reduced to 30 principal components inside each split, since 4096 features
against 240 rows is ill-conditioned. Accuracy is cross-validated over five splits, each holding out
two of the ten speakers, so no speaker appears in both training and test.

\section{Experimental setup}

\textbf{Model.} All experiments use \texttt{kyutai/moshika-pytorch-q8}, the int8 release of
Moshi~[6], with its Mimi codec at 12.5\,Hz, on a single 16\,GB RTX 4090 laptop GPU, which the
int8 checkpoint is chosen to fit. Quantization raises word error rate across all unsteered and steered generated audio, so we
report word error as a difference from the unsteered baseline on the same sentence and seed rather
than as an absolute measure.

\textbf{Vectors.} For emotion direction vector construction we use 800 clips per emotion from the English half of ESD~[7], which covers angry,
happy, sad, surprise and neutral over ten speakers, across all ten speakers.

\textbf{Audio Generation.} We use two different modes for audio generation in Moshi.

\emph{Forced text mode} fixes what is said so that only delivery can vary: all 720
Harvard sentences~[11] - the 72 phonetically balanced lists of ten from the IEEE speech-quality
standard, whose content is emotionally neutral - are forced through the inner monologue, and the
model renders each one. Any change in the emotion score is then attributable to delivery, since the
words are identical across conditions.

\emph{Free duplex mode} is the setting the model is built for and gives up that control: Moshi hears an
emotional ESD utterance on its input stream and answers in its own words, with nothing forced, 50
clips for each of five input emotions. Content is free to vary, so an emotion score there reflects
delivery and wording together, and replies are short - the model often says little. Vectors,
injection schedule and $\alpha$ are identical in both.

\textbf{Metrics and Evaluation.} emotion2vec+ large~[8] (\texttt{iic/emotion2vec\_plus\_large}) is used for emotion probability labels and emotion embeddings.
Emotion is scored only on clips with at least 1.5\,s of voiced audio, since
emotion2vec reads silence as sad and averaging over near-silent clips invents a sad result that
is not there. Every emotion number is reported against the score real ESD speech of that emotion receives from
the same classifier, which we call its \emph{ceiling}: held
out from the clips building the similarity centroids, referring to the mean classifier score and not a theoretical bound from human level speech. Similarity to real speech is reported as a mean-centred
cosine: raw cosine between two unrelated real recordings is already 0.713, so uncentred values
compress into the top quarter of their range. 

Whisper large-v3~[9] (\texttt{openai/whisper-large-v3}) is used for word error rate (WER), against the imposed sentence
under the \emph{forced text mode} and against the model's own inner monologue in \emph{free duplex mode} where no reference sentence
exists. The two therefore measure different things - fidelity to a script, and agreement between
what the model planned to say and what its codec produced - and are not directly comparable across
settings, only across conditions within one. 

\section{Results}

\subsection{Probing the Residual Stream}
\label{sec:decode}

The probe successfully reads emotion out of the residual stream at every layer (Table~\ref{tab:probe}). We report accuracy at layers 0, 8 and 24, the three
the directions are injected into, with the best of all 32 alongside. Layer 0 is weaker throughout, which is consistent with the emotion being progressively linearly separable with
depth rather than present at the input.

The contrast that will be apparent in following sections is the three-way accuracy over angry, happy and surprise - the emotions found to share highly similar
steering directions (\S\ref{sec:geom}) reaches reasonably high accuracy of 0.661 at layer 8 and 0.678 at layer 24, against a chance rate of 0.333.

The information distinguishing these emotions is therefore present in the residual stream, and the limitations in steering highlights the shortcoming to be present in the steering vectors instead of the representation.

\begin{table}[!htb]
\centering\footnotesize
\caption{Probe accuracy on the same states the directions are built from, L0, L8 and L24 are the layers the steering vectors are injected into;
the last column gives the best of all 32 layers, with that layer in brackets. \textbf{Bold} is the
strongest of the three injected layers; shaded cells sit within 30\% of the headroom above chance.}
\label{tab:probe}
\begin{tabular}{@{}lrrrrr@{}}
\toprule
contrast & chance & L0 & L8 & L24 & best (layer) \\
\midrule
angry vs happy & 0.500 & \cellcolor{cangry!12}0.567 & 0.692 & \textbf{0.750} & 0.767 (27) \\
angry vs surprise & 0.500 & 0.717 & \textbf{0.908} & 0.875 & 0.925 (11) \\
happy vs surprise & 0.500 & 0.717 & \textbf{0.758} & 0.742 & 0.792 (17) \\
angry vs sad & 0.500 & 0.767 & \textbf{0.867} & 0.850 & 0.892 (31) \\
happy vs sad & 0.500 & 0.800 & 0.892 & \textbf{0.892} & 0.933 (13) \\
sad vs surprise & 0.500 & 0.783 & \textbf{0.942} & 0.933 & 0.967 (13) \\
angry vs neutral & 0.500 & 0.683 & \textbf{0.908} & 0.875 & 0.908 (31) \\
happy vs neutral & 0.500 & \cellcolor{cangry!12}0.583 & \textbf{0.892} & 0.833 & 0.908 (9) \\
sad vs neutral & 0.500 & \cellcolor{cangry!12}0.642 & 0.758 & \textbf{0.808} & 0.850 (16) \\
surprise vs neutral & 0.500 & 0.833 & \textbf{0.925} & 0.917 & 0.958 (10) \\
\midrule
angry / happy / surprise & 0.333 & \cellcolor{cangry!12}0.517 & 0.661 & \textbf{0.678} & 0.683 (19) \\
all five & 0.200 & 0.470 & \textbf{0.603} & 0.547 & 0.607 (10) \\
\bottomrule
\end{tabular}

\end{table}

\subsection{Activation Steering}
\label{sec:steer}

Table~\ref{tab:both} displays steering accuracy for both audio generation modes together. Only sad reaches its own target, and it
overshoots: 0.650 against the 0.448 genuine ESD sad scores. Angry reaches a quarter of its ceiling
and surprise under a third, and in both the strongest rival is happy rather than the emotion asked
for.

The two settings agree closely. Angry reaches 25\% of ceiling with the forced text mode and 26\% when the
model chooses its own words; surprise 30\% and 31\%. Sad is steered exceptionally well. Word error is lower in duplex - 13.5\% for angry against 32.2\% - but reasonably so as explained in WER experimental setup section. 

\begin{table}[!htb]
\centering\footnotesize
\caption{Steering with the text forced (720 sentences per condition) and in free duplex dialogue
(50 per input emotion). \%\,c.\ is the percentage of the real-speech ceiling reached, arg\% the
share of clips whose top label is the requested emotion. Shading on a rival marks where it beats
that emotion. See Appendix~\ref{app:reading}.}
\label{tab:both}
\setlength{\tabcolsep}{3.2pt}
\begin{tabular}{@{}lrrrrlrrrlr@{}}
\toprule
& \multicolumn{2}{c}{real speech} & \multicolumn{4}{c}{forced text} & \multicolumn{4}{c}{free duplex} \\
\cmidrule(lr){2-3}\cmidrule(lr){4-7}\cmidrule(lr){8-11}
condition & P & arg\% & P(tgt) (\%\,c.) & arg\% & rival & WER & P(tgt) (\%\,c.) & arg\% & rival & WER \\
\midrule
\textit{unsteered} & 0.924 & 92.2 & \textit{0.635} & \textit{65.7} & --- & \textit{24.3} & \textit{0.505} & \textit{48.6} & --- & \textit{10.7} \\
angry steered & 0.717 & 71.9 & \cellcolor{cangry!10}0.177 (25\%) & 18.3 & \cellcolor{cangry!12}\textbf{0.328} happ & 32.2 & \cellcolor{cangry!11}0.189 (26\%) & 19.2 & \cellcolor{cangry!12}\textbf{0.370} happ & 13.5 \\
happy steered & 0.633 & 64.1 & \cellcolor{chappy!31}0.481 (76\%) & 49.3 & 0.276 surp & 61.8 & \cellcolor{chappy!22}0.331 (52\%) & 31.4 & 0.150 surp & 20.2 \\
sad steered & 0.448 & 45.3 & \cellcolor{csad!60}0.650 (145\%) & 68.6 & 0.037 happ & 51.4 & \cellcolor{csad!29}0.313 (70\%) & 32.1 & 0.199 happ & 23.5 \\
surprise steered & 0.458 & 45.3 & \cellcolor{csurprise!12}0.136 (30\%) & 13.6 & \cellcolor{cangry!12}\textbf{0.463} happ & 71.5 & \cellcolor{csurprise!13}0.143 (31\%) & 13.3 & \cellcolor{cangry!12}\textbf{0.464} happ & 35.6 \\
\bottomrule
\end{tabular}

\end{table}

We notice that for angry, surprise and happy, where the probability goes is systematic rather than diffuse. Table~\ref{tab:mass} demonstrates the full
distribution under both modes against the unsteered baseline. In both modes, we observe the audio being steered towards happy for all these three emotions.

We, additionally, provide results of steering against arbitrarily chosen random vectors as \emph{control} to reinforce the observations above. (Table~\ref{tab:control}) compares the probability scores of Neutral and Happy across four arbitrary vectors against the steered counterparts. Injecting the four arbitrary vectors leaves emotion distribution virtually unaltered while the ESD constructed steering vectors differ significantly in terms of P(emotion), and WER.

\begin{table}[!htb]
\centering\footnotesize
\caption{Four random directions injected at the same norm, through the same hooks and schedule as
the emotion directions, on 240 forced-text sentences each.}
\label{tab:control}
\begin{tabular}{@{}lrrrr@{}}
\toprule
condition & P(neutral) & P(happy) & WER\% & voiced s \\
\midrule
\textit{unsteered} & 0.599 & 0.181 & 24.7 & 1.98 \\
\addlinespace[2pt]
random 0 & 0.569 & 0.159 & 26.1 & 1.93 \\
random 1 & 0.526 & 0.197 & 21.8 & 1.98 \\
random 2 & 0.601 & 0.189 & 21.3 & 1.98 \\
random 3 & 0.537 & 0.237 & 28.7 & 1.96 \\
\addlinespace[2pt]
angry steered & 0.042 & 0.328 & 32.2 & 2.32 \\
happy steered & 0.018 & 0.481 & 61.8 & 3.04 \\
sad steered & 0.134 & 0.037 & 51.4 & 3.59 \\
surprise steered & 0.013 & 0.463 & 71.5 & 2.47 \\
\bottomrule
\end{tabular}

\end{table}

The ordering is not an end-of-clip effect. Scoring only the first 1.2\,s of each forced-text clip
(Table~\ref{tab:prefix}) already gives 0.241 happy against 0.122 angry, and 0.376 happy against
0.125 surprise.

Intelligibility degrades under both conditions, especially under \emph{forced text mode} with 71.5\% for surprise being the worst. 

\begin{table}[!htb]
\centering\footnotesize
\caption{emotion2vec probability by label: forced text on the left, free duplex on the right. Each
steered row is followed by real ESD speech of that emotion, the most any clip could score. Cell depth
is proportional to the value in that emotion's colour; \underline{underline} marks the emotion
requested and \textbf{bold} the largest in the row.}
\label{tab:mass}
\begin{minipage}[t]{0.485\linewidth}
\centering
\begin{tabular}{@{}lrrrrr@{}}
\toprule
condition & angry & happy & sad & surpr. & neutral \\
\midrule
\textit{unsteered} & \textit{0.001} & \textit{0.163} & \textit{0.035} & \textit{0.004} & \textit{0.635} \\
\addlinespace[2pt]
angry & \cellcolor{cangry!16}\underline{0.177} & \cellcolor{chappy!29}\textbf{0.328} & \cellcolor{csad!3}0.032 & \cellcolor{csurprise!7}0.084 & \cellcolor{cneutral!4}0.042 \\
\textit{real} & \textit{0.717} & \textit{0.030} & \textit{0.001} & \textit{0.001} & \textit{0.154} \\
\addlinespace[2pt]
happy & 0.007 & \cellcolor{chappy!43}\underline{\textbf{0.481}} & \cellcolor{csad!5}0.057 & \cellcolor{csurprise!24}0.276 & \cellcolor{cneutral!2}0.018 \\
\textit{real} & \textit{0.005} & \textit{0.633} & \textit{0.002} & \textit{0.006} & \textit{0.160} \\
\addlinespace[2pt]
sad & 0.008 & \cellcolor{chappy!3}0.037 & \cellcolor{csad!58}\underline{\textbf{0.650}} & 0.015 & \cellcolor{cneutral!12}0.134 \\
\textit{real} & \textit{0.001} & \textit{0.047} & \textit{0.448} & \textit{0.000} & \textit{0.392} \\
\addlinespace[2pt]
surprise & 0.002 & \cellcolor{chappy!41}\textbf{0.463} & \cellcolor{csad!16}0.186 & \cellcolor{csurprise!12}\underline{0.136} & 0.013 \\
\textit{real} & \textit{0.044} & \textit{0.177} & \textit{0.010} & \textit{0.458} & \textit{0.080} \\
\addlinespace[2pt]
\bottomrule
\end{tabular}

\end{minipage}\hfill
\begin{minipage}[t]{0.485\linewidth}
\centering
\begin{tabular}{@{}lrrrrr@{}}
\toprule
condition & angry & happy & sad & surpr. & neutral \\
\midrule
\textit{unsteered} & \textit{0.001} & \textit{0.234} & \textit{0.113} & \textit{0.025} & \textit{0.505} \\
\addlinespace[2pt]
angry & \cellcolor{cangry!17}\underline{0.189} & \cellcolor{chappy!33}\textbf{0.370} & \cellcolor{csad!2}0.027 & \cellcolor{csurprise!6}0.071 & \cellcolor{cneutral!4}0.040 \\
\textit{real} & \textit{0.717} & \textit{0.030} & \textit{0.001} & \textit{0.001} & \textit{0.154} \\
\addlinespace[2pt]
happy & 0.008 & \cellcolor{chappy!29}\underline{\textbf{0.331}} & \cellcolor{csad!13}0.146 & \cellcolor{csurprise!13}0.150 & \cellcolor{cneutral!9}0.105 \\
\textit{real} & \textit{0.005} & \textit{0.633} & \textit{0.002} & \textit{0.006} & \textit{0.160} \\
\addlinespace[2pt]
sad & 0.004 & \cellcolor{chappy!18}0.199 & \cellcolor{csad!28}\underline{\textbf{0.313}} & \cellcolor{csurprise!10}0.114 & \cellcolor{cneutral!18}0.207 \\
\textit{real} & \textit{0.001} & \textit{0.047} & \textit{0.448} & \textit{0.000} & \textit{0.392} \\
\addlinespace[2pt]
surprise & 0.003 & \cellcolor{chappy!41}\textbf{0.464} & \cellcolor{csad!12}0.134 & \cellcolor{csurprise!13}\underline{0.143} & \cellcolor{cneutral!4}0.044 \\
\textit{real} & \textit{0.044} & \textit{0.177} & \textit{0.010} & \textit{0.458} & \textit{0.080} \\
\addlinespace[2pt]
\bottomrule
\end{tabular}

\end{minipage}
\end{table}

\begin{table}[!htb]
\centering\footnotesize
\caption{Forced-text clips scored over a growing prefix rather than in full, against the strongest
rival in each window. \textbf{Bold} is the larger of the two; shading marks the windows where a rival
beats the requested emotion. $n$ falls with the window because not every clip is that long.}
\label{tab:prefix}
\begin{tabular}{@{}lrlrlrl@{}}
\toprule
& \multicolumn{2}{c}{first 1.2\,s} & \multicolumn{2}{c}{first 2.0\,s} & \multicolumn{2}{c}{whole clip} \\
\cmidrule(lr){2-3}\cmidrule(lr){4-5}\cmidrule(lr){6-7}
condition & P(tgt) & rival ($n$) & P(tgt) & rival ($n$) & P(tgt) & rival \\
\midrule
angry steered & 0.122 & \cellcolor{cangry!12}\textbf{0.241} happ (710) & 0.154 & \cellcolor{cangry!12}\textbf{0.248} happ (555) & 0.177 & \cellcolor{cangry!12}\textbf{0.328} happ \\
happy steered & \textbf{0.409} & 0.080 sad (718) & \textbf{0.433} & 0.086 surp (676) & \textbf{0.481} & 0.276 surp \\
sad steered & \textbf{0.361} & 0.084 happ (717) & \textbf{0.450} & 0.051 surp (699) & \textbf{0.650} & 0.037 happ \\
surprise steered & 0.125 & \cellcolor{cangry!12}\textbf{0.376} happ (714) & 0.104 & \cellcolor{cangry!12}\textbf{0.333} happ (600) & 0.136 & \cellcolor{cangry!12}\textbf{0.463} happ \\
\bottomrule
\end{tabular}

\end{table}

\subsection{Embedding-Similarity shows a Similar Pattern}
\label{sec:esim}

Table~\ref{tab:esim} measures the same clips a second way: mean-centred cosine to the centroid of
real ESD speech of each emotion, built from recordings disjoint from the clips scored against them.
The pattern is the same, though both readings come from emotion2vec and are therefore not independent
of one another.

\begin{table}[!htb]
\centering\footnotesize
\caption{Centred cosine to the centroid of real ESD speech of each emotion, unsteered baseline in
the first row; shading, underline and bold as in Table~\ref{tab:mass}. For angry and
surprise the nearest centroid is not the requested one.}
\label{tab:esim}
\begin{tabular}{@{}lrrrrr@{}}
\toprule
condition & angry & happy & sad & surpr. & neutral \\
\midrule
\textit{unsteered} & \textit{-0.363} & \textit{0.118} & \textit{0.305} & \textit{-0.414} & \textit{0.579} \\
\addlinespace[2pt]
angry steered & \underline{-0.023} & \cellcolor{chappy!25}\textbf{0.284} & -0.043 & -0.048 & -0.118 \\
\textit{real angry} & \textit{0.677} & \textit{-0.199} & \textit{-0.075} & \textit{-0.353} & \textit{0.011} \\
\addlinespace[2pt]
happy steered & -0.261 & \cellcolor{chappy!35}\underline{\textbf{0.394}} & -0.197 & \cellcolor{csurprise!29}0.330 & -0.339 \\
\textit{real happy} & \textit{-0.182} & \textit{0.563} & \textit{0.036} & \textit{-0.287} & \textit{0.067} \\
\addlinespace[2pt]
sad steered & -0.262 & -0.018 & \cellcolor{csad!59}\underline{\textbf{0.665}} & -0.422 & \cellcolor{cneutral!20}0.228 \\
\textit{real sad} & \textit{-0.146} & \textit{-0.137} & \textit{0.645} & \textit{-0.529} & \textit{0.382} \\
\addlinespace[2pt]
surprise steered & -0.389 & \cellcolor{chappy!34}\textbf{0.381} & \cellcolor{csad!7}0.076 & \cellcolor{csurprise!12}\underline{0.139} & -0.191 \\
\textit{real surprise} & \textit{-0.130} & \textit{0.117} & \textit{-0.235} & \textit{0.347} & \textit{-0.217} \\
\addlinespace[2pt]
\bottomrule
\end{tabular}

\end{table}

\subsection{Direction Cosine explains Off-Steering}
\label{sec:geom}

\begin{figure}[!htb]
\centering
\includegraphics[width=0.42\linewidth]{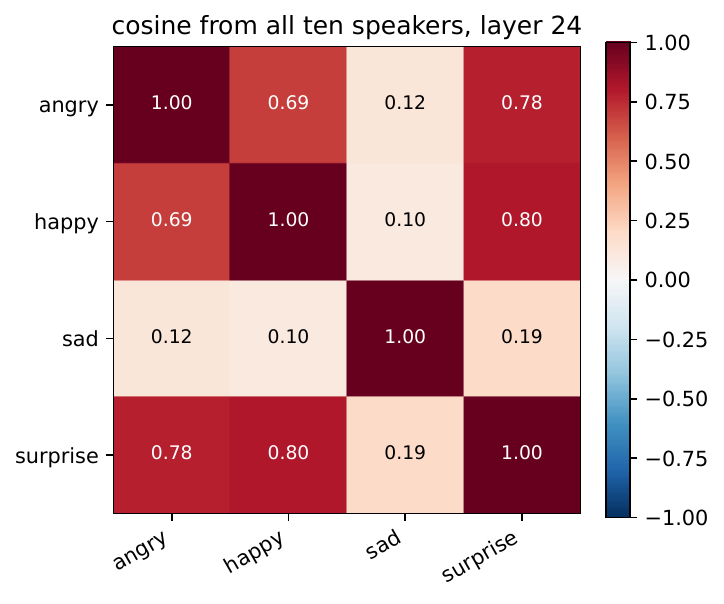}\hspace{0.02\linewidth}
\includegraphics[width=0.42\linewidth]{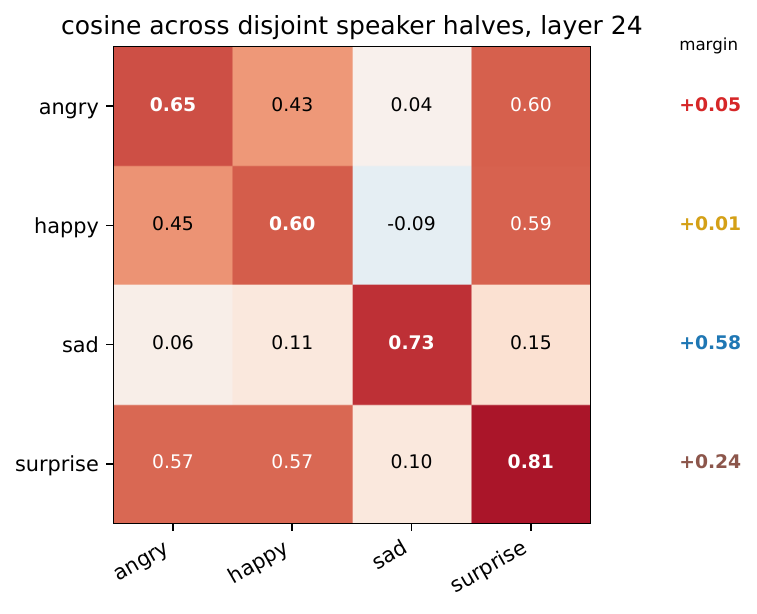}
\caption{Cosine between emotion directions at layer 24. \textbf{Left:} directions built from all ten
speakers, so each pair shares voices. \textbf{Right:} built from disjoint halves, so none do; the
bold diagonal is an emotion against itself and \emph{margin} its lead over the nearest other.}
\label{fig:geom}
\end{figure}

The steering result is reasonably explained when the steering vectors themselves are analysed
against each other. A cosine between two directions needs a scale to be read against, so we build
every direction from one half of the ten speakers and compare it with the other half's
(Figure~\ref{fig:geom}, right). 

Angry agrees with itself at 0.65 and with surprise at 0.60; happy with itself at 0.60 and with
surprise at 0.59, the margins beside each row. For these high-arousal emotions a different emotion is
nearly as close as the same emotion recorded by different speakers. Sad's nearest neighbour is 0.58
away.

Cosine across the entire corpus of speakers varies from the similarity scores on disjoint halves (Figure~\ref{fig:geom}, left). Roughly a
third of the apparent similarity between two emotion directions is shared speaker identity rather
than shared emotion - the entanglement Wang et al.~[16] report for flow-matching modules, measured
here directly.

The contrast in the margins of 0.01 to 0.05 against 0.58, for self versus cross emotion similarity, separates the emotions that reach their
own target from those that do not, and it is apparent in the direction vectors themselves prior to any audio generation.

\subsection{Removing the shared direction does not fix it completely}
\label{sec:sweep}

We may assume that if the mean difference is dominated by what these emotions share, projecting that component out
should separate them. To test this, we take the shared direction $s_l$ at layer $l$ to be the first right singular
vector of the four unit-normalised directions stacked as a $4 \times 4096$ matrix - their leading
principal component, about the origin rather than their mean since there are only four. It
carries 61--64\% of the total squared singular value at layers 0, 8 and 24. Subtracting a fraction
of each \emph{unnormalised} mean difference's component along it,
\begin{equation}
v_{e,l}(\beta) = v_{e,l} - (1 - \beta)\,(v_{e,l}^{\top} s_l)\, s_l ,
\end{equation}
leaves the plain vector at $\beta = 1$ and removes the shared component entirely at $\beta = 0$,
where angry and happy move from $+0.69$ to $-0.56$ at layer 24. Each is unit-normalised before
injection (\S\ref{sec:inject}), so every setting enters at the same magnitude. 240 sentences each.

\begin{table}[!htb]
\centering\footnotesize
\caption{Keeping a fraction $\beta$ of the direction the emotions share. Bold marks the settings
where the requested emotion beats its strongest rival. clips with
$\geq 1.5$\,s of voiced audio. The intermediate $\beta=0.25$ is in Appendix~\ref{app:sweep}.}
\label{tab:sweep}
\begin{tabular}{@{}llrrlr@{}}
\toprule
condition & $\beta$ & P(tgt) & \%\,ceil. & strongest rival & WER\% \\
\midrule
angry steered & 1.0 & {0.184} & 26 & 0.334 (happ) & 32.2 \\
 & 0.5 & \textbf{0.288} & 40 & 0.162 (happ) & 22.6 \\
 & 0.0 & {0.016} & 2 & 0.040 (happ) & 23.0 \\
\addlinespace[1pt]
happy steered & 1.0 & \textbf{0.503} & 79 & 0.279 (surp) & 62.4 \\
 & 0.5 & \textbf{0.510} & 80 & 0.175 (surp) & 65.2 \\
 & 0.0 & \textbf{0.205} & 32 & 0.115 (sad) & 73.5 \\
\addlinespace[1pt]
sad steered & 1.0 & \textbf{0.647} & 145 & 0.050 (happ) & 46.9 \\
 & 0.5 & \textbf{0.644} & 144 & 0.043 (happ) & 50.1 \\
 & 0.0 & \textbf{0.701} & 157 & 0.027 (happ) & 54.8 \\
\addlinespace[1pt]
surprise steered & 1.0 & {0.139} & 30 & 0.414 (happ) & 77.6 \\
 & 0.5 & {0.044} & 10 & 0.427 (happ) & 82.5 \\
 & 0.0 & {0.031} & 7 & 0.389 (sad) & 79.1 \\
\addlinespace[1pt]
\textit{unsteered} & --- & \textit{0.601} & --- & --- & \textit{24.7} \\
\bottomrule
\end{tabular}

\end{table}

Removing the shared direction \emph{entirely}
leaves nothing to steer with - angry falls to 0.016, below the unsteered rate - implying it carries the direction that makes angry sound angry. 

Removing \emph{half}
separates angry from happy for the first time here: P(angry) rises from 0.184 to 0.288, 26\% to 40\%
of ceiling, the happy rival falls from 0.334 to 0.162, and word error falls from 32.2\% to 22.6\%,
marginally under the 24.7\% unsteered baseline.

The sweep leaves sad almost untouched, at 0.647 to 0.701 across all four settings. That is the control
the construction provides: sad is the one direction not aligned with the shared component ($|\cos|=0.245$ at layer 24 against $0.89$--$0.94$ for the others), so the manipulation barely changes it, and the movement in angry cannot be a generic effect of perturbing the vectors. 

Surprise is
not recovered at any setting, and removal makes it worse rather than better: it falls from 0.139 to
0.031 as $\beta$ goes to zero, and never beats its strongest rival - happy at $\beta \geq 0.5$,
sad below that. Partial removal resolves one of the three merged emotions, not all of them.

\subsection{Analyzing Moshi's limitations in rendering emotions}
\label{sec:attribution}

\S\ref{sec:steer} scored steered speech against real speech, which may be a harsher baseline to compare against. Some of that loss is the model's own limitation as the model tries to propagate the stream via one forward pass. Driving a real
recording through its output codebooks and scoring the model's re-rendering of it --- the
\emph{prime} condition --- bounds how much of an emotion can reach the waveform at all through
Moshi (Table~\ref{tab:handover}).

\begin{table}[!htb]
\centering\footnotesize
\caption{Prime and release, 100 recordings per emotion, over clips clearing the 1.5\,s voiced
threshold ($n$ in brackets). \emph{gains most} is the label taking the largest share of what the
re-rendering loses. See Appendix~\ref{app:reading}.}
\label{tab:handover}
\begin{tabular}{@{}lrrrrlr@{}}
\toprule
emotion & recording ($n$) & prime ($n$) & continuation ($n$) & prime/rec.\% & gains most & silent\% \\
\midrule
angry & \cellcolor{cangry!61}0.791 (67) & \cellcolor{cangry!24}0.315 (59) & \cellcolor{cangry!6}0.073 (15) & 40 & neutral 0.06$\rightarrow$0.27 & 85 \\
happy & \cellcolor{chappy!51}0.663 (64) & \cellcolor{chappy!44}0.563 (52) & \cellcolor{chappy!18}0.234 (16) & 85 & neutral 0.03$\rightarrow$0.12 & 84 \\
sad & \cellcolor{csad!56}0.720 (68) & \cellcolor{csad!36}0.470 (63) & \cellcolor{csad!24}0.314 (15) & 65 & neutral 0.17$\rightarrow$0.33 & 85 \\
surprise & \cellcolor{csurprise!46}0.593 (75) & \cellcolor{csurprise!36}0.459 (66) & \cellcolor{csurprise!12}0.151 (14) & 77 & happy 0.08$\rightarrow$0.20 & 86 \\
\bottomrule
\end{tabular}

\end{table}

\begin{table}[!htb]
\centering\footnotesize
\caption{Steering against two references: real ESD speech, and the model's own re-rendering of it.
The \emph{real speech} column is measured differently from Table~\ref{tab:handover}'s
\emph{recording} column; Appendix~\ref{app:reading} says how.}
\label{tab:attr}
\begin{tabular}{@{}lrrrrr@{}}
\toprule
condition & P(tgt) & real speech & \%\,real & through codec & \%\,codec \\
\midrule
angry & 0.177 & 0.717 & 25 & 0.315 & 56 \\
happy & 0.481 & 0.633 & 76 & 0.563 & 85 \\
sad & 0.650 & 0.448 & 145 & 0.470 & 138 \\
surprise & 0.136 & 0.458 & 30 & 0.459 & 30 \\
\bottomrule
\end{tabular}

\end{table}

The bound is non uniform. The re-rendering keeps 85\% of the recording's happy score but only 40\% of its
angry, and the lost probability moves to \emph{neutral} instead of another emotion, implying that the
codec may be discarding the prosodic extremes to keep what it needs for intelligible reconstruction. Angry
loses most. Surprise is the only exception with a good re-render capacity of 0.459 compared to 0.593 for real recordings but a sharp drop in the continuation audio to 0.151; as it loses emotions partially towards \emph{happy}.

Scored against that bound instead (Table~\ref{tab:attr}), angry rises from 25\% to 56\% of ceiling
and happy from 76\% to 85\%: about half of angry's shortfall is the codec rather than the direction.
Surprise does not recover at all. Its bound is 0.459 against the 0.458 real surprised speech scores,
though those are different clip sets (Appendix~\ref{app:reading}); on the same clips the re-rendering
retains 77\% of the recording's surprise. Either way the codec carries most of it, so most of the gap
between 0.136 and its ceiling belongs to the direction we inject.

But much of the confusion the steering
vectors produce is already present in the shared direction and Moshi's internal incapabilities in rendering emotions, before any direction is added, leaving aside surprise which remains unexplained after above analyses.

\section{Limitations}

We test one model, one corpus, one extraction point and one injection schedule; we do not
establish that other duplex speech models behave the same way.

The int8 checkpoint puts unsteered word error at 24.3\% before any intervention, and we cannot rule
out that a bf16 model carries emotion differently. The WER is high for many steered audio samples, 
hence we offer a diagnosis rather than a comprehensive method of delivery control. Every measurement comes from emotion2vec with no
human listening test. The four emotions are ESD's, and we make no claim about how emotion is organised in general.

Modern steering algorithms such as Sparse Auto Encoders (SAEs), LoRA or full retraining/fine-tuning remain unexplored in the paper.

\section{Conclusion}

We study whether activation steering controls emotion in a full-duplex speech model, where fast and dynamic controls to delivery control is highly beneficial. We derive steering directions
from the residual stream and find that a linear probe separates the emotions in the activations while the mean-difference directions built from them barely do. For the high-arousal
emotions, two different directions agree to within a few hundredths of what a single emotion's
direction agrees with itself across a change of speaker. Steering follows that geometry rather than the requested label, and only the emotion
whose direction stands apart is controlled as its label implies. The result holds with the text
forced as well as in a full duplex dialogue. We propose a generation-free check that explains it from the
vectors alone, and show that the shared component is necessary rather than incidental: removing it
entirely leaves a direction that steers nothing, while retaining half of it restores one of the merged
emotions. 
Modern estimators built to discriminate smartly rather than to average, and the same comparison applied
to a model finetuned for controllability, fall under further extensions to this work.

\clearpage
\appendix

\section{Reading the tables}
\label{app:reading}

\paragraph{Which clips a number covers.} Emotion scores are taken over clips with at least
1.5\,s of voiced audio, since emotion2vec reads silence as sad. Table~\ref{tab:excluded} reports how
many that removes from each condition.

\paragraph{Two different real-speech references.} Table~\ref{tab:handover}'s \emph{recording}
column is the 100 clips per emotion that were actually primed, under the voiced filter.
Table~\ref{tab:attr}'s \emph{real speech} column is a separate set of 64 clips per emotion, held out
from those used to build the similarity centroids and scored without the filter. They are therefore
not the same measurement, and differ most for sad and surprise, whose clips are quietest: 0.720
against 0.448 for sad, 0.593 against 0.458 for surprise. The second is the ceiling every
``\%\,of ceiling'' figure in the paper is taken against, so those percentages compare filtered
generated speech with unfiltered real speech.

\paragraph{Prime and release.} Each recording yields three clips: the original, the model's
re-rendering of it while its output codebooks are overridden, and its free continuation once the
override is lifted. \emph{silent} is the percentage of continuations falling below the voiced
threshold, so that column rests on a small and non-random subset and is not a retention rate.

\paragraph{Cosine panels.} In Figure~\ref{fig:geom} (right) every direction is built from one half
of the ten speakers and compared with the other half's, so no two directions being compared share a
voice. The left panel builds both directions from all ten. The off-diagonal falls by roughly a third
between the panels, and that difference is speaker identity rather than emotion.

\section{Clips excluded by the voiced-audio threshold}
\label{app:excluded}

emotion2vec scores silence as sad, so every emotion number in the paper is taken over clips with at
least 1.5\,s of voiced audio. Table~\ref{tab:excluded} reports what that removes.

The filter is not neutral between conditions. Under forced text it drops 8.1\% of unsteered clips
but only 0.3--1.4\% of steered ones, because steering lengthens the utterance - voiced duration
rises from 1.91\,s unsteered to 2.30--3.58\,s. The unsteered baseline that every comparison rests on
is therefore measured over a differently filtered subset than the conditions it is compared with. In
duplex the threshold removes 58--71\% of clips in every condition, which is why its numbers are
noisier than the forced-text ones.

\begin{table}[h]
\centering\footnotesize
\caption{Clips reaching the 1.5\,s voiced-audio threshold.}
\label{tab:excluded}
\begin{tabular}{@{}llrrr@{}}
\toprule
run & condition & kept & total & dropped\% \\
\midrule
forced text & unsteered & 662 & 720 & 8.1 \\
forced text & angry & 710 & 720 & 1.4 \\
forced text & happy & 718 & 720 & 0.3 \\
forced text & sad & 717 & 720 & 0.4 \\
forced text & surprise & 714 & 720 & 0.8 \\
duplex & unsteered & 72 & 250 & 71.2 \\
duplex & angry & 73 & 250 & 70.8 \\
duplex & happy & 105 & 250 & 58.0 \\
duplex & sad & 81 & 250 & 67.6 \\
duplex & surprise & 75 & 250 & 70.0 \\
\bottomrule
\end{tabular}

\end{table}

\section{Every setting of the shared-direction sweep}
\label{app:sweep}

Table~\ref{tab:sweepfull} adds the $\beta = 0.25$ setting omitted from Table~\ref{tab:sweep}. It sits
where the trend predicts and changes no conclusion: angry at 0.119 is already past its optimum on the
way down to 0.016, and no emotion's ordering against its strongest rival differs from the neighbouring
settings.

\begin{table}[h]
\centering\footnotesize
\caption{All four settings of $\beta$.}
\label{tab:sweepfull}
\begin{tabular}{@{}llrrlr@{}}
\toprule
condition & $\beta$ & P(tgt) & \%\,ceil. & strongest rival & WER\% \\
\midrule
angry steered & 1.0 & {0.184} & 26 & 0.334 (happ) & 32.2 \\
 & 0.5 & \textbf{0.288} & 40 & 0.162 (happ) & 22.6 \\
 & 0.25 & \textbf{0.119} & 17 & 0.081 (happ) & 15.3 \\
 & 0.0 & {0.016} & 2 & 0.040 (happ) & 23.0 \\
\addlinespace[1pt]
happy steered & 1.0 & \textbf{0.503} & 79 & 0.279 (surp) & 62.4 \\
 & 0.5 & \textbf{0.510} & 80 & 0.175 (surp) & 65.2 \\
 & 0.25 & \textbf{0.534} & 84 & 0.092 (sad) & 68.7 \\
 & 0.0 & \textbf{0.205} & 32 & 0.115 (sad) & 73.5 \\
\addlinespace[1pt]
sad steered & 1.0 & \textbf{0.647} & 145 & 0.050 (happ) & 46.9 \\
 & 0.5 & \textbf{0.644} & 144 & 0.043 (happ) & 50.1 \\
 & 0.25 & \textbf{0.658} & 147 & 0.034 (happ) & 49.9 \\
 & 0.0 & \textbf{0.701} & 157 & 0.027 (happ) & 54.8 \\
\addlinespace[1pt]
surprise steered & 1.0 & {0.139} & 30 & 0.414 (happ) & 77.6 \\
 & 0.5 & {0.044} & 10 & 0.427 (happ) & 82.5 \\
 & 0.25 & {0.043} & 9 & 0.385 (sad) & 84.1 \\
 & 0.0 & {0.031} & 7 & 0.389 (sad) & 79.1 \\
\addlinespace[1pt]
\textit{unsteered} & --- & \textit{0.601} & --- & --- & \textit{24.7} \\
\bottomrule
\end{tabular}

\end{table}

\section{Implementation}
\label{app:code}

Both listings are the code as run.

\subsection{Injection}

A per-layer tensor of shape $[\text{batch}, 4096]$ is added to that layer's output; zeroing a row
steers nothing, which is how the unsteered condition shares a batch with the steered ones. The
direction is unit-normalised and scaled by the layer's own median residual norm, so $\alpha$ means
the same thing at every depth.

\begin{lstlisting}
def install_steering(lm, batch_size, vector_dir=None):
    deltas = {layer: torch.zeros(batch_size, 4096, device="cuda") for layer in INJECT_LAYERS}
    for layer in INJECT_LAYERS:
        lm.transformer.layers[layer].register_forward_hook(
            (lambda l: (lambda mod, inp, out: out + deltas[l][:, None, :].to(out.dtype)))(layer))

    source = vector_dir or GENERATED / "vectors"
    layer_norms = json.load(open(source / "layer_norms.json"))
    vectors = {path.stem[4:]: np.load(path) for path in sorted(source.glob("vec_*.npy"))}

    def make_delta(emotion, layer, alpha=0.30):
        direction = vectors[emotion][layer]
        direction = direction / np.linalg.norm(direction)
        scale = alpha * max(0.0, 1 - layer / 31) * layer_norms[str(layer)]
        return torch.tensor(scale * direction, device="cuda")

    return deltas, make_delta
\end{lstlisting}

\subsection{Reading the state}

The recording is written into the model's own output codebooks in place of what the depformer
would emit, while the transcript is forced through the inner monologue. Forward hooks record the
residual stream, and the last few frames before the override ends are averaged to give one state
per clip.

\begin{lstlisting}
def _prime_one_batch(mimi, lm, tok, clips, waveforms, n_frames, layers, keep, continuation,
                     per_frame=False):
    batch = len(clips)
    recorded = {layer: [] for layer in layers}
    handles = [lm.transformer.layers[layer].register_forward_hook(
        (lambda l: (lambda mod, inp, out: recorded[l].append(out[:, 0].float().cpu())))(layer))
        for layer in layers]

    padded = torch.zeros(batch, 1, n_frames * FRAME, device="cuda")
    for slot, wav in enumerate(waveforms):
        padded[slot, 0, :len(wav)] = wav.cuda()
    codes = mimi.encode(padded)[:, :8]

    set_seed(4242)
    force, _ = force_text(lm, tok, [c["text"] for c in clips], batch)
    step = [0]

    def on_text(tokens):
        if step[0] < n_frames:
            force(tokens)
        step[0] += 1

    silence = lambda t: torch.zeros(batch, 1, FRAME, device="cuda")
    audio, text = generate(mimi, lm, batch, n_frames + continuation, silence, on_text=on_text,
                           replace_output=lambda t: codes[:, :, t:t + 1] if t < n_frames else None)
    for handle in handles:
        handle.remove()

    if per_frame:
        stacked = {layer: torch.stack(recorded[layer]) for layer in layers}
        offset = stacked[layers[0]].shape[0] - len(text)
        assert offset >= 0, (stacked[layers[0]].shape[0], len(text))
        return [(np.stack([stacked[layer][offset:, slot].numpy() for layer in layers],
                          axis=1).astype(np.float16), text[:, slot])
                for slot in range(batch)]

    calls = len(recorded[layers[0]])
    window = {layer: torch.stack(recorded[layer][calls - keep:calls]) for layer in layers}
    return [(np.stack([window[layer][:, slot].numpy().mean(0) for layer in layers]),
             audio[slot, 0][:n_frames * FRAME],
             audio[slot, 0][n_frames * FRAME:],
             waveforms[slot].numpy())
            for slot in range(batch)]
\end{lstlisting}

\subsection{The probe}

Fitted per layer on the states the directions are built from. \texttt{groups} holds a speaker-pair
index, so \texttt{LeaveOneGroupOut} never places a speaker in both training and test, and the scaler
and projection are refitted inside each fold.

\begin{lstlisting}
def probe_accuracy(features, target, groups):
    def pipeline(n_train):
        components = min(30, n_train - 1)
        return make_pipeline(StandardScaler(), PCA(components, random_state=0),
                             LogisticRegression(max_iter=2000))
    scores = [pipeline(len(train)).fit(features[train], target[train]).score(
        features[test], target[test])
        for train, test in LeaveOneGroupOut().split(features, target, groups)]
    return float(np.mean(scores))
\end{lstlisting}

\subsection{Removing the shared direction}

The shared direction is the first right singular vector of the four unit-normalised directions at
each layer; \texttt{beta} is how much of each emotion's component along it the swept vector keeps.

\begin{lstlisting}
unit = lambda v: v / (np.linalg.norm(v) + 1e-9)
neutral = states[emotion == "neutral"].mean(0)
plain = {e: states[emotion == e].mean(0) - neutral for e in EMOTIONS}

shared_direction = {}
for index, layer in enumerate(layers):
    stacked = np.stack([unit(plain[e][index]) for e in EMOTIONS])
    _, singular, right = np.linalg.svd(stacked, full_matrices=False)
    shared_direction[index] = right[0]
    share = singular[0] ** 2 / (singular ** 2).sum()
    alignment = {e: round(float(unit(plain[e][index]) @ right[0]), 3) for e in EMOTIONS}
    print(f"layer {layer:<3} shared direction explains {100 * share:.1f}% of the four; "
          f"cosine to it {alignment}")

for beta in RETENTION:
    for e in EMOTIONS:
        vector = np.zeros((32, 4096), np.float32)
        for index, layer in enumerate(layers):
            direction = plain[e][index]
            along = float(direction @ shared_direction[index])
            residue = direction - along * shared_direction[index]
            vector[layer] = residue + beta * along * shared_direction[index]
        np.save(out_dir / f"vec_{e}_b{beta}.npy", vector)
\end{lstlisting}

\end{document}